%% file: bare_jrnl_transmag.tex
\documentclass[journal,transmag]{IEEEtran}
\usepackage{amssymb}
\usepackage{mathtools}

\usepackage{array}
\usepackage{amsmath}
\usepackage{algorithm}
\usepackage{algpseudocode}
\usepackage{xcolor}
\usepackage{subfigure}
\usepackage{multirow}
\usepackage{cite}

\ifCLASSINFOpdf
\else
\fi
\begin{document}
%
\title{Inter-Domain Fusion for Enhanced Intrusion Detection in Power Systems: An Evidence Theoretic and Meta-Heuristic Approach}


\author{\IEEEauthorblockN{Abhijeet Sahu\IEEEauthorrefmark{1},~\IEEEmembership{Student Member,~IEEE}
Katherine Davis\IEEEauthorrefmark{1},~\IEEEmembership{Senior Member,~IEEE}}
\IEEEauthorblockA{\IEEEauthorrefmark{1}Electrical and Computer Engineering,
Texas A\&M University, College Station, TX 77843 USA}
\thanks{ 
Corresponding author: A. Sahu (email: abhijeet\_ntpc@tamu.edu).}}

%



\IEEEtitleabstractindextext{%
\begin{abstract}
False alerts due to misconfigured or compromised intrusion detection systems (IDS) in industrial control system (ICS) networks can lead to severe economic and operational damage. 
To solve this problem, research has focused on leveraging deep learning techniques that would help reduce false alerts.
However, a shortcoming is that these works often require or implicitly assume the physical and cyber sensor data to be trustworthy.  Implicit trust of data is a major problem with using artificial intelligence or machine learning (AI/ML) for cyber-physical system (CPS) security, because the times when these solutions are needed most to detect an attack are also the times when they are more at risk, with both greater likelihood and greater impact, of also being compromised. 
To address this inevitable shortcoming, the problem can thus be reframed as how to make good decisions given uncertainty. Then, the decision is detection, and the uncertainty includes whether or not the data that would be used in ML-based IDS is compromised.
Thus, this article presents an approach for reducing false alerts in cyber-physical power systems that addresses this critical problem of dealing with uncertainty without the knowledge of prior distribution of the alerts.
Specifically, an evidence theoretic based approach leveraging Dempster Shafer (DS) combination rules and their variants is proposed for reducing false alerts.  A multi-hypothesis mass function model is designed that leverages probability scores obtained from various supervised-learning classifiers.  Using this model, a location-cum-domain based fusion framework is proposed to evaluate the intrusion detector’s performance using Disjunctive, Conjunctive and Cautious Conjunctive rules of combinations, that fuse multiple piece of evidences from inter-domain and intra-domain sensors. The approach is demonstrated in a cyber-physical power system testbed (RESLab), and the classifiers are trained with datasets from Man-In-The-Middle attack emulation in a large-scale synthetic electric grid.  For evaluating the performance, we consider plausibility, belief, pignistic, general Bayesian theorem based metrics as decision functions. 
To improve the performance, a multi-objective based genetic algorithm is proposed for feature selection considering the decision metrics as the fitness function.
Finally, we present a software application to evaluate the DS fusion approaches with different parameters and architectures.
\end{abstract}

}

\maketitle

\IEEEdisplaynontitleabstractindextext

%
\IEEEpeerreviewmaketitle


\input{Contents/1_introduction}

\input{Contents/2_background}
\input{Contents/3_dst}

\input{Contents/3_5_ga_prob}

\input{Contents/4_testbed_model}

\input{Contents/5_fusion_architecture}


\input{Contents/7_results}

\input{Contents/8_application}

\input{Contents/9_conclusion}


%

\appendices
\section{Types of mass function}\label{app_mf}
\begin{enumerate}
    \item {Normal: If $\O$ is not a focal set or $m($\O$) = 0$}
    \item {Subnormal: If $\O$ is a focal set.} In our experiments, the null set in the focal set hence subnormal belief function is considered.
    \item {Dogmatic: If $\Omega$ is not a focal set. In the experiments, $\Omega$ is in the focal set hence it has nondogmatic belief function. The combine cautious rule can only be implemented if the mass distribution is nondogmatic. }
    \item {Vacuous: If $\Omega$ is the only focal set.} For certain timestamp when there are no alerts, we also observe vacuous focal set.
    \item {Simple: If it has at most two focal sets and, if it has two, $\Omega$ is one of them.}
    \item {Categorical: If it has only one focal set.}
    \item{Bayesian: If its focal sets are singleton.}
\end{enumerate}

\section{DS Functions}\label{app_df}

\subsection{Belief Function}
The belief function maps each hypotheses B to a value bel(B) between 0 and 1, defined in in Eq.~\ref{eq:2}:
\begin{equation}\label{eq:2}
    \begin{aligned}
        bel(B) = \sum_{j:A_j \subset B} m(A_j)
    \end{aligned}   
    \vspace{-2mm}
\end{equation}
In words, belief in a hypothesis B is the sum of masses of elements which are subsets of A, $A_j$.

\subsection{Plausibility Function}
The plausibility function maps each hypotheses B to a value pls(B) between 0 and 1, defined in Eq.~\ref{eq:2b}:
\begin{equation}\label{eq:2b}
    \begin{aligned}
        pls(B) = \sum_{j:A_j \cap B \ne \O} m(A_j)
    \end{aligned}    
    \vspace{-2mm}
\end{equation}
In words, pls(B) is the sum of all the mass of the sets, $A_j$ that intersects with the set B.
It has an alternate definition, i.e., the weight of evidence that does not refute B, and hence the belief and plausibility are related by Eq.\ref{eq:3}.
\begin{equation}\label{eq:3}
    \begin{aligned}
        pls(B) = 1 - bel(\overline{B})
    \end{aligned}   
    \vspace{-2mm}
\end{equation}
\subsection{Commonality Function}
The commonality function maps each hypotheses B to a value q(B) between 0 and 1, defined in Eq.~\ref{eq:4}:
\begin{equation}\label{eq:4}
    \begin{aligned}
        q(B) = \sum_{j:B \cap A_j} m(A_j)
    \end{aligned} 
    \vspace{-2mm}
\end{equation}

\subsection{Relationship among the function}
Shafer showed that there is one-to-one correspondence among these m, bel, pls and q functions, where Eqs.~\ref{eq:5} and \ref{eq:6} give examples:
\begin{equation}\label{eq:5}
    \begin{aligned}
        bel(B) = \sum_{j:A_j \supseteq B} (-1)^{|A_j|-|B|} q(A_j), \forall B \subseteq \Omega
    \end{aligned}  
    \vspace{-2mm}
\end{equation}

\begin{equation}\label{eq:6}
    \begin{aligned}
        m(B) = \sum_{j:A_j \subseteq B} (-1)^{|B|-|A_j|} bel(A_j), \forall B \subseteq \Omega
    \end{aligned}   
    \vspace{-2mm}
\end{equation}


\section*{Acknowledgment}
The work described in this paper was supported by funds from the National Academy of Sciences under award PO\# 2000009323 titled \textit{Online Resilience Support System for Cyber-Physical Situational Awareness} and from US Department of Energy's (DoE) \textit{Cybersecurity for Energy Delivery Systems program} under award DE-OE0000895.

\bibliographystyle{IEEEtran}
\bibliography{myreference}

\ifCLASSOPTIONcaptionsoff
  \newpage
\fi

\end{document}

%% file: Contents/1_introduction.tex
\section{Introduction}
\label{intro}


%
The increase of advanced control and communication technologies within the electric power grid can 
make the system more vulnerable to cyber intrusions. 
Several ICS-targeted attacks such as Stuxnet ~\cite{stuxnet}, Ukraine~\cite{ukraine}, Mumbai~\cite{mumbai} are well known for severe impacts with advanced concept of operations. The criticality of power grid infrastructure necessitates the design of resilient detection and defense mechanisms to prevent such attacks.

The behavior of cyber intrusions and their impact on a network is stochastic in nature. This \textit{stochasticity} is typically modeled using Markov Processes, where the transition probabilities depend on attributes represented graphically such as the degree of the nodes and the prior distribution of the states of the nodes~\cite{stochastic}. Similarly, \textit{uncertainty} is an innate feature of any intrusion analysis. The uncertainty arises due to the defender’s 
inability to completely view the adversary’s steps, as the monitoring tools can only observe certain symptoms or effects of malicious activities. 

Stochasticity and uncertainty complicate cyber intrusion detection and incident forensics. Intrusion Detection Systems (IDS) commonly rely on rule-based policies (\textit{signature-based}) or deviations from a baseline (\textit{behavioral-based}) to detect cyber intrusions. These systems produce false alarms, both false negatives and false positives. Signature-based IDSs result in higher false negatives for stealthy~\cite{achilles} and zero-day attacks.  Behavioral-based IDSs, while based on statistics, often result in high false positives.  High false positives are detrimental to an organization’s efficiency and effectiveness at threat response because they cost time and money for security professionals to investigate, and they erode an organization’s trust in the system’s results. False negatives also pose a significant threat, since an undetected attack may escalate more privileges to result in increased damage or loss to the organization’s assets. For Industrial Control Systems (ICS), IDSs may be further customized based on process-data analysis, control-command analysis, and with help of an ICS physical model~\cite{ics_ids_survey}. 
While security tools such as IDSs and firewalls provide key functions, they are typically assumed to be trustworthy. Furthermore, obtaining the data needed for theoretical models to improve the function of such security tools is a challenge, as even behavioral-based IDSs do not have enough intrusion information to build the statistical models~\cite{challenges_ids}. The lack of trust in the IDS creates uncertainty in the evidences from sensors.

To address these challenges due to stochasticity, uncertainty, and lack of adequate data availability,
we present a cyber-physical power system intrusion detection system based on the theory of uncertainty, which we call the \textbf{I}nter-\textbf{D}omain \textbf{E}vidence theoretic \textbf{A}pproach for \textbf{I}nference in cyber-physical power systems (IDEA-I).  We address the problem of high false alarms in IDS, through the solution we develop that works to leverage fusion of evidence by domain and location using \textit{Dempster-Shafer (DS)} rules of combination. IDEA-I is based on an \textit{autonomous} data fusion architecture~\cite{hall_book}, where the features extracted are fed to the classifiers or estimators for decision making before they are fused.  This is \textit{decision-level} fusion, where each sensor performs individual processing to produce an estimate, and then these estimates are combined in the fusion process.  There are numerous methods possible to achieve the fusion process, such as voting methods, Bayesian inference, DS methods, and generalized evidence processing theory~\cite{hall_book}. 
DS inference~\cite{zadeh} is a fusion technique applicable to the autonomous fusion architecture, and so is Bayesian~\cite{intro1} inference, because these fusion algorithms are fed with the probability distributions computed from the classifiers or the estimators. 

In IDEA-I, we propose the usage of Dempster-Shafer Theory of Evidence (DSTE) for network detection in power system control networks. This approach provides value in how it handles uncertainty due to its ability to quantify unknowns.  Specifically, two advantages we achieve from D-S theory are (1) its ability to deal with the lack of prior probabilities for various events and (2) its ability to combine evidences from multiple sources~\cite{zomlot}.

The major contributions of this paper are as follows:
\begin{enumerate}
    \item A cyber-physical power system intrusion detection system IDEA-I is proposed that improves intrusion detection by inferring cyber-physical state information to improve situational awareness based on the fundamentals of DSTE, various rules of fusion and decision criteria.
    \item A method for computing mass functions for 
    stochastic
    cyber-physical parameters, 
    from the detection probability computed in our prior work on data fusion~\cite{paper3},
    is proposed 
    and evaluated in IDEA-I. 
    The performance based on two different architectures, location and location-cum-domain based fusion, using IDEA-I is evaluated. 
    \item IDEA-I is extended to formulate a feature selection unconstrained optimization problem and solved using Non-dominating Sorted Genetic Algorithm (NSGA)~\cite{nsga2}, to improve IDEA-I accuracy. 
    \item IDEA-I is developed as a software tool that includes the development of a DSTE library in C\#. The application is used to evaluate the performance of the proposed fusion algorithm for varying scenarios and parameters.
    
\end{enumerate}

The paper is organized as follows.  Section~\ref{sec:background} presents background on the DSTE approach.  Section~\ref{dst} describes IDEA-I including how the method would need to work in cyber-physical power systems with its rules of combinations and their implications. A Genetic Algorithm (GA)-based optimization problem is proposed in Section~\ref{sec:ga} for feature selection to improve IDEA-I performance. Section~\ref{sec:testbed} presents the experimental setup in our cyber-physical power system testbed RESLab, the use cases that were designed to test IDEA-I, and their implementation.  Then, Section~\ref{sec:fusionarchitecture} introduces the two types of architecture proposed for the fusion.  We compare the approach and results with the centralized-based fusion and other decision-level fusions such as Bayesian inference.  The overall results are discussed in Section ~\ref{sec:results}, and Section~\ref{sec:conclusion} concludes the paper.

%% file: Contents/2_background.tex
\section{Background}\label{sec:background}

DSTE is applied in many areas of machine learning and deep learning. An unsupervised classification problem in multisource remote sensing is formulated through DS theory, as one can consider union of classes rather than individual class~\cite{ds_app0}. 
A neural network based classifier is proposed where the DS computation of mass function or basic belief assessment (BBA) and rules of combination are implemented in two hidden layers respectively~\cite{ds_app1}. 
Majority of the research are either centered towards wireless networks~\cite{ds_app15}, network security~\cite{ds_app11}, or autonomous mobile robots~\cite{ds_app2}. 

DSTE is also being applied to network security. In~\cite{ds_app9}, multi-source alarm information is fused through DSTE which is associated with nodes vulnerability information, integrated with the severity of threats for situational assessment of network security. A network anomaly detector with enhanced reliability with low false alarms is proposed using DSTE~\cite{ds_app10}. An IDS is proposed in~\cite{ds_app11} where the mass function are computed based on the incoming and outgoing traffic ratio, service rate and the prior knowledge in the domain of DDoS attacks. A distributive and collaborative based IDS is proposed using DSTE for fusion data from multiple nodes~\cite{ds_app12} where the detection is done collaboratively and the decision is distributed among all nodes.

The presented work, IDEA-I, is the first to leverage DS theory for the purpose of classification based on the dataset~\cite{paper3} generated from MiTM attacks in a cyber-physical power system testbed~\cite{Sahu2020}.

Data fusion in a cyber-physical system 
should utilize data from the physical (e.g., power) and cyber sensors as evidences to generate belief functions for the hypothesis. Examples include root vulnerability exploitation or situational awareness for attack prevention. Cyber-physical frameworks for situational awareness~\cite{davis,sahu_fw,socca} have been proposed that identify critical assets and contingencies using power system simulators, graph theories, dynamic programming, etc. For example, one framework~\cite{davis} builds a partially observable Markov Decision Process (POMDP) model of the grid network that represents all possible attack paths. The robustness of such a framework primarily depends on fusion of information like network access policies, firewall rulesets, physical sensors, etc. The transition probabilities of the security states in the POMDP model depend on the amount of data accrued in real-time. However, uncertainty is present due to unavailability of complete view of the adversary's steps and monitoring limitations.
%
%
The presented IDEA-I addresses this gap through its use of DSTE for power system cyber-physical situational awareness, that handles uncertainty due to its ability of quantifying unknowns. 
Analogous to the space situational awareness (SSA) paradigm~\cite{marcus}, we improve the cyber-physical situational awareness (CyPSA) framework by accurately representing the state knowledge of objects in the cyber-physical environment to provide better prediction capabilities for potential threats.

DSTE suffers from major drawbacks of its computational requirements and the challenges it encounters while eliciting the probability masses from multiple evidence~\cite{dste_ga}. Hence, to address this, we have proposed the use of formulating an optimization problem, by taking the decision function from DSTE as the objective function, for feature selection to train the classifier. Since there are multiple decision metrics, we employ a multi-objective optimization problem and solve using a meta-heuristic GA approach. GA has been used extensively in network intrusion detection such as flow-based traffic characterization~\cite{ga_ref1}, IDS rules generation~\cite{ga_ref2}, feature selection~\cite{ga_ref3}, etc. GA in the DSTE framework was proposed in the turbine maintenance optimization problem~\cite{dste_ga}. In this work, we present GA with DSTE 
in cyber-physical security for feature selection.

%% file: Contents/3_dst.tex
\section{Development of IDEA-I from Dempster Shafer Theory \&  Combination Rules}
\label{dst}

Uncertainty is classified into two categories based on knowledge and behavior of the system: Aleatory and Epistemic uncertainty~\cite{dste_ga}. \textit{Aleatory uncertainty} is caused due to random behavior of system, while \textit{Epistemic uncertainty} is caused due to lack of knowledge of the system. Under normal operation of a SCADA or OT network, the traffic are not random as in an IT network, hence aleatory uncertainty rarely occurs. Under a compromised situation, the system state is ignorant rather than stochastic, hence the uncertainty in events is epistemic. For example, a zero-day attack cannot be detected by knowledge-based or signature-based IDS, due to lack of information about the intrusions. Dempster and Shafer introduced the belief function for modeling epistemic uncertainty for reasoning under uncertainty. Quantifying uncertainty with a precise measure is difficult, and hence a measure of probability as an interval is considered. Three major frameworks for interval-based representation of uncertainty are the following: a) Imprecise probability, b) Possibility Theory, and  c) Demspter Shafer Theory of Evidence (DSTE). DSTE is preferred because of its high degree of theoretical development, better relationship with traditional probability theory, large engineering applications in the past few years, and the versatility of the theory to represent and combine different types of evidences. 

In evidence theory, logs at each sensor act as evidence that are considered for reasoning of an event. Theoretically, there are four types of evidence: a) Consonant, b) Consistent, c) Arbitrary, and d) Disjoint~\cite{evidence_type}. 
The data of a control network with an IDS for cyber intrusion detection and a bad data detector for power system state estimation can be collectively considered as Disjoint evidence due to their different purpose of deployment. 
Traditional probability theory cannot handle consonant, consistent, or arbitrary evidence without resorting to assumptions in distributions. DSTE can handle all these kinds by combining a notion of probability with the traditional conception of sets.

\begin{figure}[h]
  \centering
\includegraphics[width=0.9\linewidth]{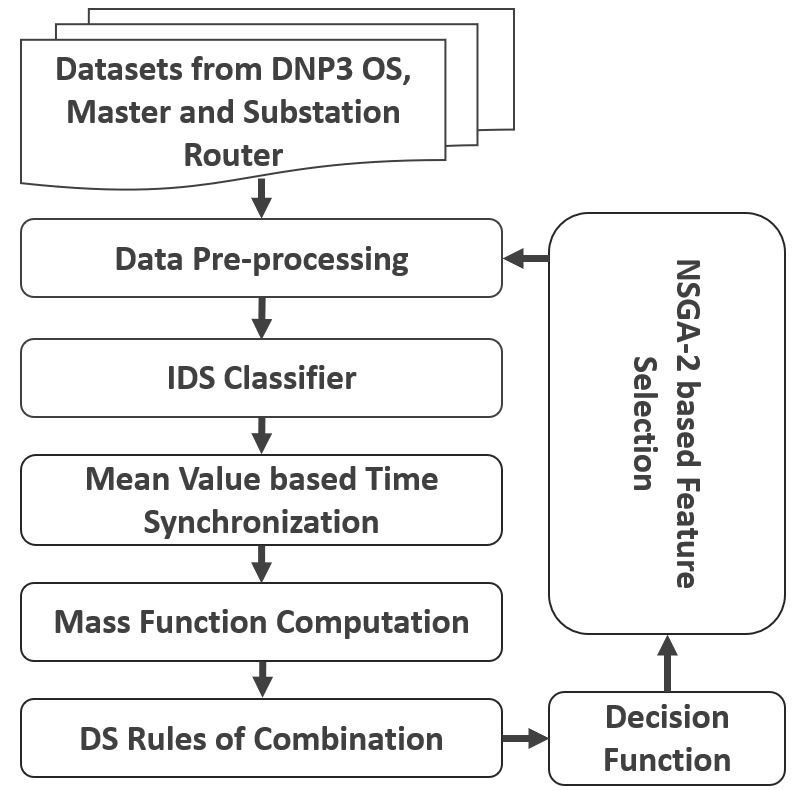}
  \caption{Sequence of operations, with feedback from DSTE decision function for optimal feature selection using meta-heuristic algorithm. }
  \label{fig:flow_chart}
\end{figure}

The IDEA-I framework is illustrated in Fig.~\ref{fig:flow_chart}. The \textit{Datasets} are sensor data extracted from substation networking devices, DNP3 Master and Outstations from the RESLab testbed~\cite{Sahu2020}. DNP3~\cite{clarke2004practical} is a protocol used in SCADA systems for monitoring and controlling field devices.
\textit{Data Pre-processing} is performed before training the ML based \textit{IDS Classifier}. The output from the classifier from different sensors carries data of varying timestamps which synchronized with the \textit{Mean Value based Time Synchronization} block. The calculations for \textit{Mass Function Computation}, \textit{DS Rules of Combination}, and the \textit{Decision Function} blocks 
of DSTE are detailed in this section. The decision function is used in the fitness function in the version 2 of NSGA \textit{i.e. NSGA-2 based Feature Selection} block (Section~\ref{sec:ga}) to again filter the features in the Data Pre-processing block. Each block in the flow-chart is detailed further. 

\subsection{Dempster Shafer Theory of Evidence (DSTE)}
Fundamentally, DSTE can be separated from the basic probability theory on the basis of the manner one distributes the probability density or mass based on the type of random variables. For example, probability theory assigns $0.5$ to both $Head$ and $Tail$ for the toss of an unbiased coin. However, DS theory assigns a $0$ belief to $Head$ and $Tail$ but assigns a $1$ belief to the set $\{Head, Tail\}$, i.e., \textit{``Either Head or Tail."} DS does not compel picking a probability when there is no evidence. This approach provides three kinds of answers: $Yes$, $No$, and \textit{Don't Know}. Allowing the third option, i.e., \textit{ignorance}, can make evidential reasoning valuable
when there are not enough data to validate a hypothesis.

DSTE is concerned with \textit{bounds} for probabilities of provability, rather than computing probabilities of truth. The two bounds are called \textit{belief} and \textit{plausibility}. Equivalent to the state space in probability, there is a set of mutually exclusive and exhaustive hypotheses denoted by $\Omega$, also called the \textit{Frame of Discernment}. The set of all possible subsets of $\Omega$ , including itself and the null set $\O$, is called a power set and designated by $2^{\Omega}$. Thus, the power set comprises all possible hypotheses or so-called \textit{focal elements}.

\subsection{Basic Belief Assignment in DSTE}
The basic belief assignment (BBA) function or the mass distribution function ($m$), distributes the belief over the power set of the frame of discernment. Subsets $A$ of $\Omega$ such that $m(A) > 0$ are called \textit{focal sets} of $m$. This mass distribution function can be classified into $Normal$, $Subnormal$, $Dogmatic$, $Vacuous$, $Simple$, $Categorical$, and $Bayesian$. The definitions of each are presented in Appendix~\ref{app_mf}.

A subnormal BBA can be transformed to a normal BBA $m^*$ by the normalization operation defined in Eq.~\ref{eq:1}:

\begin{equation}\label{eq:1}
    m^{*}(A) = 
    \begin{cases}
    k.m(A) & if A \ne 0\\
    0 & otherwise
    \end{cases}
\end{equation}

\noindent However, the utmost care is required before normalization: the authors in~\cite{zadeh} explain a controversial issue in the normalization of the upper and lower probabilities and its implication on the rules of combination of the evidences. 
The different functions considered in the Decision Function block of DSTE (Fig.~\ref{fig:flow_chart}) for validating a hypothesis are presented in Appendix~\ref{app_df}.

\subsection{Rules of Combination}
The purpose of aggregation of information is to summarize a collection of data, whether the data is coming from a single source or from multiple sources.  

\subsubsection{Dempster's Rules of Combination (DRC)}
Dempster's rules of combination is a procedure for combining independent piece of evidence. The requirement for establishing the independence of sources is an important philosophical question. From a set theoretic standpoint, these rules can potentially occupy a continuum between \textit{Conjunction} (AND-based on set intersection) and \textit{disjunction} (OR-based on set union)~\cite{prade}. For a situation where all the evidences are reliable, a \textit{conjunctive} operation is appropriate, while for one reliable source, \textit{disjunctive} operation is preferred. Hence, in the domain of intrusion detection, one cannot rely on all the IDSs and network logs to give reliable information due to the existence of false positives and negatives in the algorithms. Thus, in these scenarios, one should prefer the disjunctive rule.


However, there are many other combination operations such as \textit{A and B or C}, \textit{A and C or B}, etc. Prade~\cite{prade} describes these three types of combination as conjunctive pooling ($A \cap B$, if $A \cap B \ne \O$), disjunctive pooling ($A \cup B$), and a tradeoff of both of them.
The original combination rule of multiple basic probability assignments known as \textit{the Dempster rule} is a generalization of \textit{Bayes’ rule}. This rule strongly emphasizes the agreement between multiple sources and ignores all the conflicting evidence through a normalization factor. The mathematical operation $\oplus$, corresponds to the normalized conjunctive fusion rule:

\begin{equation}\label{eq:7}
    \begin{aligned}
        m_{1,2}(A) = (m_1 \oplus m_2)(A) \\= \frac{1}{1-K}\sum_{B \cap C = A \neq \emptyset} m_1(B)*m_2(C)
    \end{aligned}    
\end{equation}
where \begin{equation}
    K = \sum_{B \cap C = \emptyset} m_1(B)*m_2(C)
\end{equation}

The disjunctive rule of combination is given by:
\begin{equation}\label{eq:7b}
    m_{1,2}(A)=\left\{\begin{array}{ll}
\sum_{A=B \cup C } & m_{1}\left(B\right) m_{2}\left(C\right), \quad A \neq \emptyset 
\\
0, & \quad \quad\quad\quad \quad\quad\quad\quad A=\emptyset
\end{array}\right.  
\end{equation}

\subsubsection{Combine Cautious (CC)}
The Combine Cautious rules of combination is based on the work~\cite{combine_cautious}. Conventional DS rules of combination require the evidence from multiple source to be distinct or independent, which may not be true in a realistic application. Many works have developed mechanisms to overcome the limitations of the distinctness assumptions, but they were limited to at most two focal sets. These methods were extended to separable belief functions, but since all belief functions are not separable, the conventional method was not further extended. The operators in these rules of combination, need to satisfy the mathematical properties such as \textit{associative}, \textit{commutative}, and \textit{idempotency}. Many rules of combination that were developed either did not obey those requirements, or they were not scalable for large focal sets. Moreover, the conjunctive rule is based on the assumption that the belief functions to be combined are induced from reliable sources of information. Due to the above challenges in the DRC, the CC method of combination is considered.

In the DRC, combination rules belong to the \textit{credal level} where the evidence are aggregated, but the decisions rules are implemented at the \textit{pignistic level}~\cite{tbm}.
In the CC, the weight function are computed using the commonality function defined in Appendix~\ref{app_df} given by the following,
\begin{equation}
    \begin{aligned}
w(A) &=\prod_{B \supseteq A} q(B)^{(-1)^{|B|-|A|+1}} \\
&=\left\{\begin{array}{ll}
\frac{\prod_{B \supseteq A,|B| \notin 2 \mathbb{N}} q(B)}{\prod_{B \supseteq A,|B| \in 2 \mathbb{N}} q(B)} & \text { if }|A| \in 2 \mathbb{N} \\
\frac{\prod_{B \supseteq A,|B| \in 2 \mathbb{N}} q(B)}{\prod_{B \supseteq A,|B| \notin 2 N} q(B)} & \text { otherwise }
\end{array}\right.
\end{aligned}
\end{equation}

\noindent where $2N$ denotes the set of even natural numbers.

As per the Least Commitment Principle used in combine cautious~\cite{combine_cautious}, the rules of combination are as follows:
Let $m_1$ and $m_2$ be two consonant BBA.  The $m$ is said to be consonant, if its focal sets are nested i.e. $p l(A \cup B)=p l(A) \vee p l(B), \quad \forall A, B \subseteq \Omega$, where $\vee$ is the maximum operator), and let $q_1$ and $q_2$ be their respective commonality functions. Then, the consonant BBA $m_{12}$ with commonality function $q_{12}(A) = q_{1}(A) \wedge q_{2}(A) \forall A \subseteq \Omega$ is claimed to be the s-least committed element in the set $\mathcal{S}_{S}\left(m_{1}\right) \cap \mathcal{S}_{s}\left(m_{2}\right)$, where $S$ is the specialization matrix. Hence, the cautious combination rule for the two nondogmatic BBAs is given by $w_{12}(A) = w_{1}(A) \wedge w_{2}(A), \forall A \subset \Omega$, where $\wedge$ is the minimum operator and each $w$ is the respective weight function.

\subsection{Decision Criteria}\label{sec:decisioncriteria}
\subsubsection{Belief and Plausibility Scores} After the fused mass function are computed using disjunctive, conjunctive and cautious combine rules, the belief and plausibility scores are calculated using Eqns.~\ref{eq:2} and ~\ref{eq:2b}, in Appendix B.
\subsubsection{Pignistic Scores}
To take a rational
decision, we propose to transform beliefs into pignistic probability functions through the generalized pignistic transformation (GPT)~\cite{tbm}. The pignistic transformation is based on the following equation,
\begin{equation}
P\{A\}=\sum_{X \in 2^{\Theta}} \frac{|X \cap A|}{|X|} m(X)
\end{equation}
where $|A|$ is the number of the worlds present in the set $A$, and $X$ are the other components in the frame of discernment. Usually decisions are made by computing the expectation over multiple simulations, using the pignistic $P\{.\}$ as the probability function needed to compute expectations. Usually, one uses the maximum of the pignistic probability as decision criterion. The max of $P\{.\}$ is often considered as a prudent betting decision criterion between the two other alternatives such as max of plausibility or belief function.

\subsubsection{General Bayesian Theorem (GBT)}

The GBT is a generalization of Bayes' theorem, except the conditional probabilities in Eq.~\ref{eq:bt}
are replaced by belief functions, and the \textit{a priori} belief function on
$\Theta$ is vacuous. GBT can be used for backward propagation of the belief networks to compute the posterior probabilities induced on $\Theta$ for any $x \in X$.

\begin{equation}\label{eq:bt}
    P\left(\theta_{i} | x\right)=\frac{P\left(x | \theta_{i}\right) P_{0}\left(\theta_{i}\right)}{\sum_{j} P\left(x | \theta_{j}\right) P_{0}\left(\theta_{j}\right)} \quad \forall \theta_{i} \in \Theta
\end{equation}


%% file: Contents/3_5_ga_prob.tex
\section{Genetic Algorithm for Feature Selection}
\label{sec:ga}
Feature selection  is a challenging task, for intrusion detection in a cyber physical system involves uncertainty in intrusion events. The complexity of the CPS model increases when the cyber and physical network models are defined in detail.  Leveraging feature reduction techniques such as PCA can assist in improving detection accuracy, but the feature transformation can result in un-identification of the decipherable features. Hence, it is crucial to adopt non-transformable techniques in feature reduction using optimization techniques while also considering system uncertainty. 

%
The feature selection problem based on the stochastic system, which relies on the epistemically uncertain parameters, can be formulated as a multi-objective optimization problem with uncertain objective functions (i.e., the belief, plausibility, pignistic scores of the hypothesis). In this context, the objective of the present work is to propose a feature selection technique by propagating the uncertainties of the conventional classifiers onto the fitness values and formulating the solution of the GA as a binary encoding. GA was previously used in network intrusion detection such as flow-based traffic characterization~\cite{ga_ref1}, IDS rules generation~\cite{ga_ref2}, feature selection~\cite{ga_ref3},  etc. 

In this work, a GA-based meta-heuristic approach is adopted for feature selection,
and the initial population consists of chromosomes of randomly selected features. The detection probabilities of these randomly selected features are computed by training them through the classifiers. Then, the fitness functions (belief, plausibility, and pignistic) are obtained for the different evidences, which were further used for selection, mutation and cross-over operation. Since there are multiple hypothesis in this problem, a multiple objective problem is formulated. 


\subsection{NSGA-2}
A single fitness function cannot provide an optimal solution for the multiple decision metrics considered in the DSTE framework. Hence, multi-objective GA algorithms need to be explored. NSGA~\cite{nsga2} has been found to solve multi-objective problems efficiently. In this paper, a faster version of NSGA (NSGA-2) has been adopted to solve the feature selection problem. 


The algorithm for NSGA-2 is given in Algorithm~\ref{alg:nsga2}. It involves primarily two steps: \textbf{a)} From the given population, $P_t$, at iteration $t$, the offspring solution, $Q_t$, is obtained using the selection, mutation, and crossover operations (Line 12-15). In the first step, using the union of $P_t$ and $Q_t$, non-dominated sorting is performed to obtain solutions at different pareto-front levels (Line 2-3). Non-dominates sorting is a sorting done between two solutions, say $X$ and $Y$, where $X$ is considered to dominate $Y$, if and only if there is no objective of $X$ worse than that objective of $Y$ and there is at least one objective of $X$ better than that objective of $Y$. Pareto-Front is a set of non-dominated solutions, being chosen as optimal if no objective can be improved without sacrificing at least one other objective. \textbf{b)} In the second step, while the next population set $P_{t+1}$ is obtained by sequentially adding the elements in the obtained pareto fronts, starting with 1 until the condition $|P_{t+1}| + |F_i| \le N$ is satisfied (where $F_i$ is the solution in the $i^{th}$ front, and $N$ is the maximum size of the population), for the selection of the elements in $F_i$, crowding-distance computation using the fitness function in each front (Line 6) is performed to obtain diverse solutions (Line 5-9).

\begin{algorithm}
\caption{Algorithm of NSGA-2}\label{alg:nsga2}
\begin{algorithmic}[1]

\While{termination criteria}
\State $R_t \gets P_t \cup Q_t$
\State $F \gets $ non\_dominated\_sorting($R_t$)
\State $P_{t+1} \gets \phi ; i \gets 1$
\While {$|P_{t+1}| + |F_i| \le N$}
 \State $C_{i} \gets $ crowd\_sourcing\_assignment($F_i$)
 \State $P_{t+1} \gets P_t \cup F_i$
 \State $i = i + 1$
\EndWhile
\State $F_i \gets sort(F_i, C_i, desc)$
\State $P_{t+1} \gets P_{t+1} \cup F_i[1:(N - |P_{t+1}|)]$
\State $Q_{t+1} \gets selection(P_{t+1},N)$
\State $Q_{t+1} \gets mutation(Q_{t+1}) $
\State $Q_{t+1} \gets crossover(Q_{t+1})$
\State $t \gets t+1$
\EndWhile
\end{algorithmic}
\end{algorithm}

\subsection{Problem Formulation}
The objective of the problem is to minimize the error with reference to the attack labels $A(t)$, over the sampled time throughout the simulation, so as to identify the least number of features that need to be considered for training the classifiers,
\begin{equation} \label{f1}
        \min \quad F_{k}=\sum_{t=0}^{N_{sim}} |f_{k}(t) - A(t)|
        \vspace{-2mm}
\end{equation}
where $k \in K$, $K$ is the set of all decision metrics such as fused Belief, Plausibility, Pignistic, GBT functions, $f_{k}$ is their corresponding scores after the fusion operations, as presented in Section~\ref{sec:decisioncriteria} and \ref{dst}, respectively, and $N_{sim}$ is the simulation duration. The decision variables are binary encoded indicating whether a feature is selected for training the classifier or not.
The $f_{k}(t)$ at time $t$ depends on the feature selected for training. $A(t) \in {0,1}$ depending on the attack window, i.e., $A(t) = 1$ during attack or else $A(t)=0$.

%% file: Contents/4_testbed_model.tex

\section{Testbed \& Fusion Architecture 
}\label{sec:testbed}

Before discussing the location-based fusion, it is essential to understand the architecture of the RESLab testbed that is producing the data during emulation of different Man-in-the-Middle attacks. 
\subsection{Testbed Architecture}

The RESLab emulation testbed consists of a network emulator, a power system emulator, an OpenDNP3 master and a RTAC based master, an intrusion detection system, and data storage, fusion and visualization software, shown in Fig.~\ref{fig:testbed}. A brief overview of each component is given below. The detailed explanation of RESLab, including its architecture and use cases, is provided in~\cite{Sahu2020,paper2}.
Common Open Research Emulator (CORE) is used to emulate the communication network. PowerWorld Dynamic Studio (PWDS) is a real-time simulation engine for operating the simulated power system case in real-time as a server~\cite{powerworld}. DNP3 Masters are incorporated using open DNP3. Snort is used in the testbed as the rule-based, open-source intrusion detection system (IDS). The Elasticsearch, Logstash, and Kibana (ELK) stack is used to probe and store all virtual and physical network interface traffic. A separate VM is deployed to operate the fusion engine that collects network logs and Snort alerts from the ELK stack using the Elasticsearch client and raw packet captures from CORE using pyshark. The fusion engine constructs cyber and physical features and merges them, using the time stamps from different sources to ensure correct alignment of information. Further, it pre-processes the features using imputation, scaling, and encoding before training for intrusion detection using supervised, unsupervised and semi-supervised learning techniques. More details can be found in~\cite{paper3}.

\begin{figure}[h]
  \centering
\includegraphics[width=1.0\linewidth]{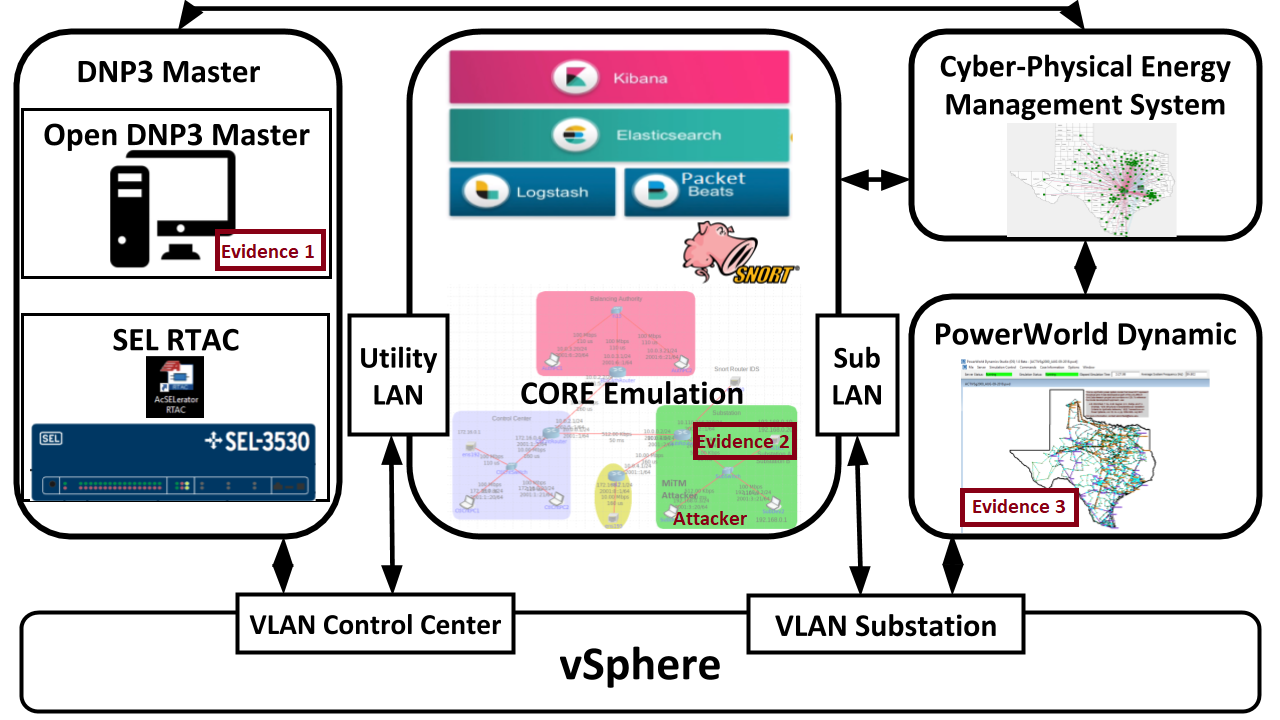}
  \caption{RESLab Emulation Testbed Architecture showing three evidences in three different locations: DNP3 Master (Evidence 1), Substation Router (Evidence 2) and PowerWorld DS acting as DNP3 outstation (Evidence 3).}
  \label{fig:testbed}
\end{figure}

\subsection{Modifying Measurements and Commands}
\label{usecases}
The objective of the intruder is to 
disrupt grid operations. 
Details on the sequence of actions that create the FCI and FDI attacks and how they impact the power system side are detailed in
\cite{Sahu2020,paper2}. These are four use cases:

\textit{Use Case 1 ($UC\;1$): Branch Control Modifications. \:} 
Each binary direct operate command is changed from a \textit{CLOSE} to a \textit{TRIP} command, with any other traffic simply forwarded. The change in the binary operate command introduces some processing delay, which may cause the packet to be retransmitted.

\textit{2. Use Case 2 ($UC\;2$): Generator Set-Point Modification. \:}
When the MiTM script is running, the analog point for the generator is set to a lower value, which decreases the generator setpoint from its current value.

\textit{3. Use Case 3 ($UC\;3$): Measurement and Status Modification. \:} $UC\;3$ is a combination of false command and data injection attacks. After each polling interval, the DNP3 master will send a read request packet to each outstation, which then sends a read response packet to the master. This read response is filled with the all the binary input, analog input, binary output, and analog output DNP3 points. Next, analog input points in the read response packet are changed to a lower value lower of 20~MW or 0~MW. The operator controlling the DNP3 master is then forced to send an analog direct operate command to bring the generators back to their original loaded set points. However, when the operator sends this original set point value to the generator, the MiTM script is programmed to change the setpoint to 20~MW or 0~MW.

\textit{4. Use Case 4 ($UC\;4$): Measurement and Status Modification. \:}
The adversary first follows the steps of Use Case~3, then modifies the read response packet of the preceding packets, based on the actual set point given by the master. Thus the master is unaware of the contingency created.

\subsection{Use Cases Implementation}
For each use case, the polling intervals and the number of polled DNP3 outstations are altered. The polling intervals tested were 30 and 60~s, while the number of polled DNP3 outstations were five and ten. For instance, the scenario $UC1\;10\;OS\; 30$ indicates $Use\;Case\;1$ with ten outstations and a polling interval of 30~s is implemented. 

In each scenario, the normal operation is conducted first without the MiTM attack. Then the operation is conducted again with the attack to analyze its impact. Finally, the attack is stopped and the network restored.
The main reason for choosing poll intervals of 30 and 60~s is that most DNP3 masters have polling rates of 30~s, 1~min, or 5~min, with a maximum of 15~min.
A poll interval more than two minutes has little impact on attack strength
because the adversary processing time is less than 60 to 70~ms.
Similarly, outstation numbers of five and ten are considered, since the objective is to study the communication dynamics of an impacted outstation, 
and how the number of outstations becomes a limitation on the
attack success probability. The numbers of five and ten 
coincide with our use cases in the Texas 2000-bus synthetic model, where each utility control center is communicating  with at least two substations and at most 25 substations.




\subsection{Mean Value Based Time Synchronization}
The classifier probability scores of detection and the time stamp may vary for different locations. The sample times will also vary. Hence, a time resolution window $res$ is selected to compute the average of the probability scores from samples existing in that window and store the average probability score.  This ensures time synchronization for fusion by location. The lower the window size, the higher the time resolution will be, but more noise will be present in the decision function. The impact of resolution is studied considering accuracy of the fusion technique for $res=5,10,15,20s$.

\subsection{Supervised learning based IDS}
Different supervised learning based classifier are used in the Data Fusion Engine~\cite{paper3}. The probability scores based on the classifier's output for each data point are considered for computing the mass function from each evidence. In the testbed, the IDSs are trained at three different locations in the network: a) DNP3 outstation, b) DNP3 master, and the c) substation router. Hence, adopting the autonomous fusion architecture is appropriate, which is more decision-centric fusion. Then location based fusion is considered using the DS rules of combinations. Initially, seven types of supervised-based classifiers are trained: a) Support Vector Classifier (SVC), b) K-Nearest Neighbor (kNN), c) Decision Tree (DT), d) Random Forest (RF), e) Gaussian Naive Bayes (GNB), f) Bernoulli Naive Bayes (BNB), and g) Multi-layer Perceptron (MLP) to compute the probability scores for different use cases with varying poll rates and polled number of outstations.  Then, the value of the decision function is used to evaluate the belief mass to feed into the DS-based fusion engine. Further, the mass distribution is computed based on the probability score. The frame of discernment for the given IDS problem is given by~${\{attack\};\{no\_attack\};\{attack,no\_attack\};\{\emptyset\}}$. If the probability of intrusion for a data point $t$ is say $a$, then the dogmatic belief mass distribution is set to be the following,
\begin{equation}
\begin{multlined}
    m_t(\emptyset) = -1\sum_{i=1}^{2} {(m_t(i) - 1/2)}^{2}\\
    m_t(attack) = a*(1-m_t(\emptyset))\\
    m_t(no\_attack) = (1-a)*(1-m_t(\emptyset))\\
\end{multlined}
\end{equation}
where the first mass belief distribution $m_t(\emptyset)$ quantifies uncertainty, as per the variance of the probability scores considered in~\cite{ds_classifier}. However, since most if not all states of belief are based on imperfect and not entirely conclusive evidence, non-dogmatic belief functions should be considered where $m(\Omega)$ is very small~\cite{combine_cautious}, say $\epsilon$  and not zero. In this scenario, a different belief mass distribution is proposed:
\begin{equation}
\begin{multlined}
    m_t(\emptyset) = -1\sum_{i=1}^{2} {(m_t(i) - (1-\epsilon)/2)}^{2}\\
    m_t(attack) = a*(1-m_t(\emptyset))*(1-\epsilon)\\
    m_t(no\_attack) = (1-a)*(1-m_t(\emptyset))*(1-\epsilon)\\
    m_t(attack,no\_attack) = m(\Omega) = \epsilon*(1-m_t(\emptyset))\\
\end{multlined}
\end{equation}

\subsection{Fusion Architectures}\label{sec:fusionarchitecture}

\begin{figure*}[t]
  \centering
\includegraphics[width=1.0\linewidth]{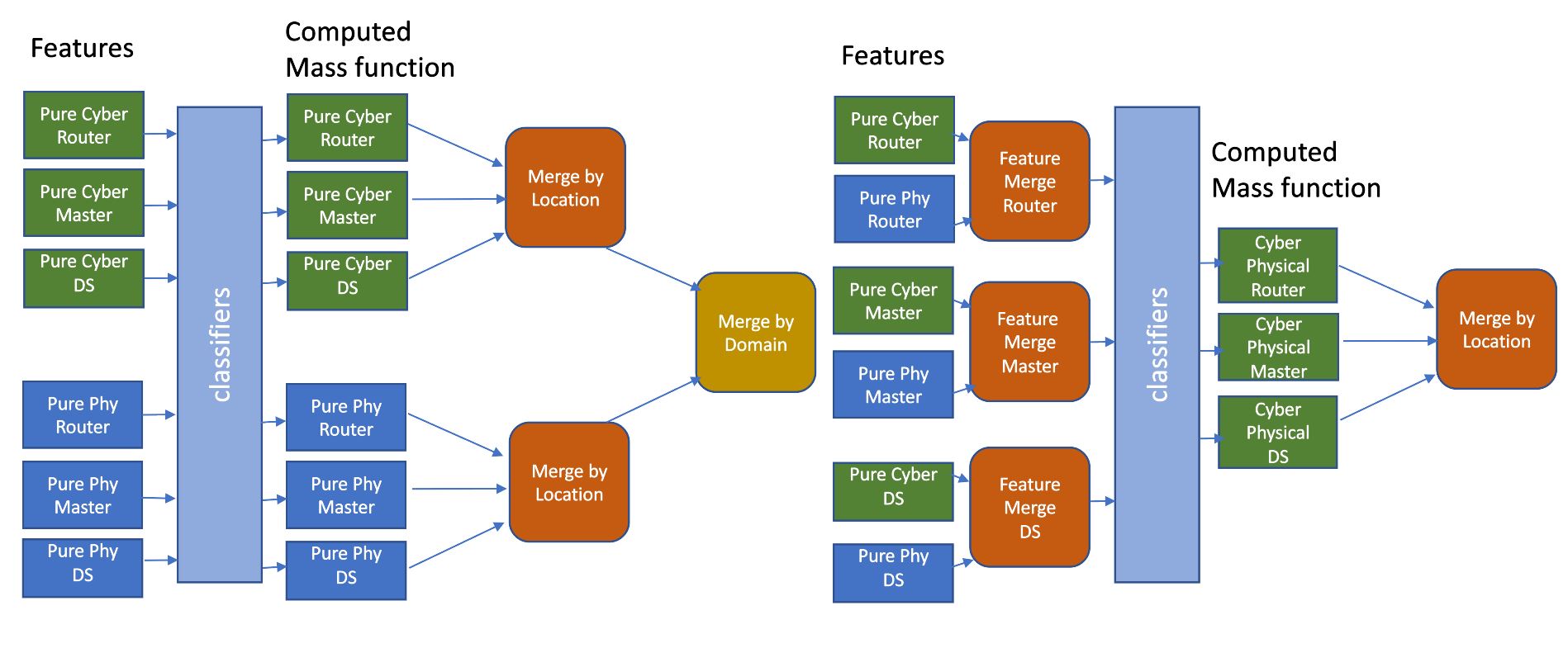}
\vspace{-1cm}
  \caption{(Left) Fusion by location and domain from pure cyber and physical classifiers. (Right) Fusion by location from cyber-physical classifiers.}
  \label{fig:two_architecture}
\end{figure*}

Two types of fusion architectures are proposed and experimented as shown in Fig.~\ref{fig:two_architecture}.
\subsubsection{Fusion by Location (FL)}
In the first case, we explore the performance through fusion by location based on the probability scores obtained from classifiers trained with both cyber and physical features at the substation router, DNP3 master, and outstation. 
\subsubsection{Fusion by Location and Domain (FLD)}
In the second case, we fuse by location as well as domain, utilizing the probability scores obtained from classifiers trained with pure cyber and physical features. 

%% file: Contents/5_fusion_architecture.tex




%% file: Contents/7_results.tex
\section{Results and Analysis} \label{reslab_exp}\label{sec:results}

This section describes the experiments performed on evaluating the performance of IDEA-I on the data collected from the sensors in the RESLab testbed~\cite{Sahu2020}, where different Man-In-The-Middle attacks were implemented in the emulated synthetic electric grid. Different rules of combination and decision criteria are evaluated from the DSTE. Further, these criteria or scores were used to compare different classifiers that must be considered prior to incorporation of DS rules of combination. 
Post combination, the scores are fed to the NSGA-2 algorithm for feature selection to improve the performance. 
Based on the selected classifier, different types of fusion techniques used in DSTE are evaluated. The performance of two architecture introduced in the previous section that involves fusion by location and domain are assessed.  Finally, the impact of the time resolution in the fusion operation on its accuracy alongwith the time complexity of the fusion algorithm is analysed with varying number of hypothesis in the belief mass distribution.    

\subsection{Decision Criteria Selection}
\begin{figure}[h]
\centering
  \includegraphics[width=0.9\linewidth]{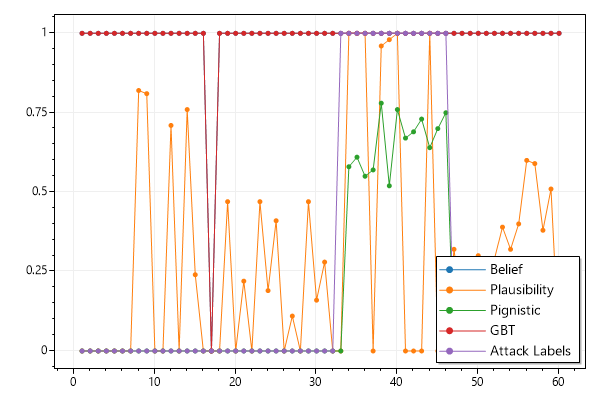}
  \caption{Plot of the decision criteria after distinctive fusion, with the probability scores from decision tree classifiers for use case 1 with 10 polled outstation and 30s poll interval}
  \label{fig:decision_score_curve}
  \vspace{-4mm}
\end{figure}

In DSTE, different criteria or scores serve different purposes. 
Hence, four decision criteria a) Belief score, b) Plausibility score, c) Pignistic score, and d) Score based on General Bayesian Theorem (GBT) presented earlier are evaluated. The basic idea is to evaluate the accuracy under these different criteria, for all the use cases, while considering different classifiers and cyber-physical features, and finally to select the criteria that has the highest accuracy in the most scenarios. For the evaluations, the time resolution $res$ is assumed to be 15~s, and the disjunctive rule of combination is considered. The accuracy is calculated based on $\frac{TP+TN}{TP+TN+FP+FN}$, where $TP$ and $TN$ are true positives and negatives, $FP$ and $FN$ are false positives and negatives. It can be observed from Table~\ref{tab:comparison_criteria} that for seven classifiers and 14 use cases, the \textit{Pignistic score} has the highest accuracy under 78 scenarios. The \textit{Belief score}, \textit{Plausibility score}, and \textit{GBT score} have the highest accuracy under 7, 7, and 8 scenarios respectively. Thus, results show the pignistic score as a reliable criteria for evaluation.  Fig.~\ref{fig:decision_score_curve} shows the decision metrics for the disjunctive fusion performed on the evidence collected from router, master, and outstation IDS with a decision tree classifier. The pignistic score is a better indicator for intrusion detection compared to the other scores. These scores are also utilized to formulate the fitness function for the meta-heuristic based feature selection and to re-evaluate the DS rules of fusion. 

\begin{table}[h]
    \begin{tabular}{|l|l|l|l|l|l|l|l|l|l|}
     \cline{1-10}
\multicolumn{3}{|c|}{Scenarios} & \multicolumn{7}{|c|}{Classifiers}\\
    \cline{1-10}
    UC & os & PI & SVC &K-NN & DT & RF & GNB & BNB & MLP\\
    \cline{1-10}
        \multirow{2}{*}{\textbf{UC1}}  & 10 & 30 & c & c & c & c & b & c & b\\ \cline{2-10}
        & 10 & 60 & c & c & c & c & c & c & c\\ \cline{1-10}
        \multirow{4}{*}{\textbf{UC2}}  & 5 & 30 & c & c & c & c & c & c & c\\ \cline{2-10}
       & 5 & 60 & c & c & d & c & c & c & c\\ \cline{2-10}
        & 10 & 30 & c & c & a & c & c & c & c\\ \cline{2-10}
       & 10 & 60 & c & c & a & c & b & c & c\\ \cline{1-10}
       \multirow{4}{*}{\textbf{UC3}}  & 5 & 30 & c & c & c & c & c & c & c\\ \cline{2-10}
       & 5 & 60 & c & c & a & c & c & c & c\\ \cline{2-10}
       & 10 & 30 & b & c & c & c & c & a & c\\ \cline{2-10}
       & 10 & 60 & b & c & c & c & c & b & c\\ \cline{1-10}
       \multirow{4}{*}{\textbf{UC4}}  & 5 & 30 & c & c & a & c & d & b & d\\ \cline{2-10}
       & 5 & 60 & c & c & a & c & c & c & d\\ \cline{2-10}
       & 10 & 30 & c & c & d & c & d & d & d\\ \cline{2-10}
       & 10 & 60 & c & c & c & c & c & c & a\\ \cline{1-10}
       \hline
    \end{tabular}
    \caption{Decision criteria with maximum accuracy evaluated accross different scenarios and classifiers (a = Belief score, b= Plausibility score, c= Pignistic score; d= GBT score).}
    \label{tab:comparison_criteria}
    \vspace{-4mm}
\end{table}

\subsection{Comparison by Classifier}
\begin{figure*}[h]
  \centering
  \subfigure[]{\includegraphics[height=1.2 in,width=1.75 in]{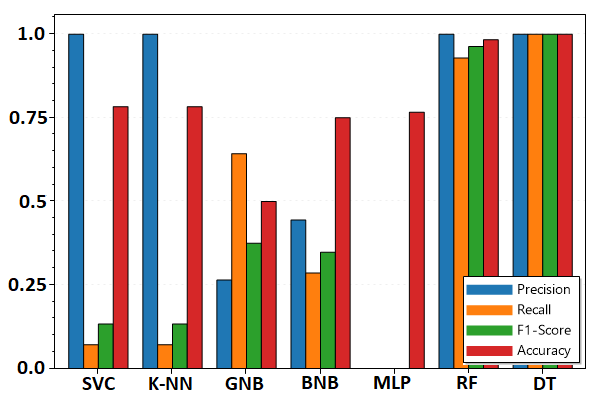}\label{uc1}}
  \subfigure[]{\includegraphics[height=1.2 in,width=1.75 in]{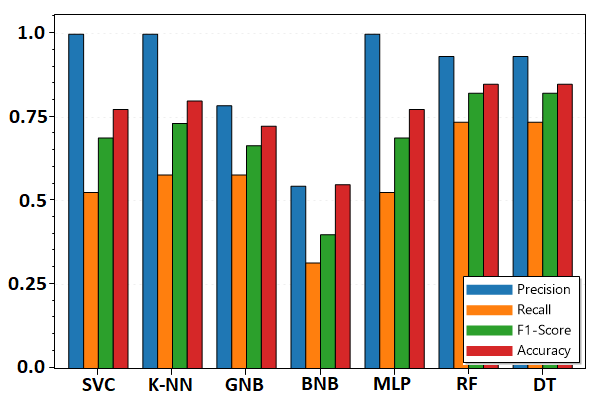}\label{uc2}}
  \subfigure[]{\includegraphics[height=1.2 in,width=1.75 in]{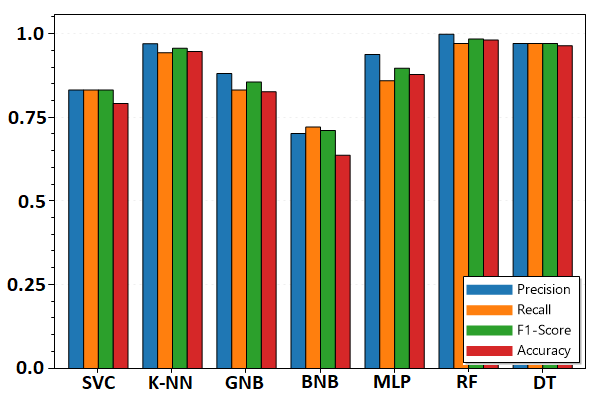}\label{uc3}}
  \subfigure[]{\includegraphics[height=1.2 in,width=1.75 in]{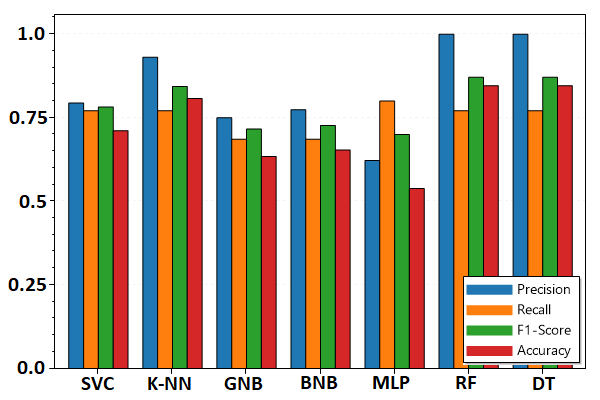}\label{uc4}}
  \caption{Precision, Recall, F1 score, Accuracy obtained based on the pignistic score from disjunctive based DS fusion performed from probability scores from different classifiers with cyber-physical features combined using \textbf{FL architecture} in Fig.~\ref{fig:two_architecture} for (a) UC1; (b) UC2; (c) UC3; (d) UC4 with 30~sec polling interval and 10 polled outstations.}
  \vspace{-4mm}
\end{figure*}

A classifier's performance may vary based on the features selected for training the learning engine. Certain use-cases perform well with the use of linear classifiers such as SGD or logistic regression, while some outperform with the use of non-linear classifiers such as DT, RF, or SVM. Some cases perform better with the use of deep learning based classifiers such as CNN or RNN, depending on the spatial and temporal nature of features and relationship with the labels.
Here, the seven classifiers are compared with the $FL$ architecture of fusion. Figs.~\ref{uc1},~\ref{uc2},~\ref{uc3},~\ref{uc4} show the comparison of the classifiers used, prior to the disjunctive based fusion, evaluated based on the precision, recall, F1-score and accuracy calculated after disjunctive based location fusion. For UC-1, UC-2, UC-3, and UC-4, the DT- and RF-based classifiers are found to be better options with disjunctive-based fusion. However, the precision scores 
for UC-1 and UC-2 were better with SVM and k-NN classifiers, but accuracy and F1-score are considered the major criteria here, so DT- and RF-based classifiers are considered for further experiments. 

\subsection{Comparison of Rules of Combination}
\begin{figure*}[h]
  \centering
  \subfigure[]{\includegraphics[height=1.6 in,width=2.3 in]{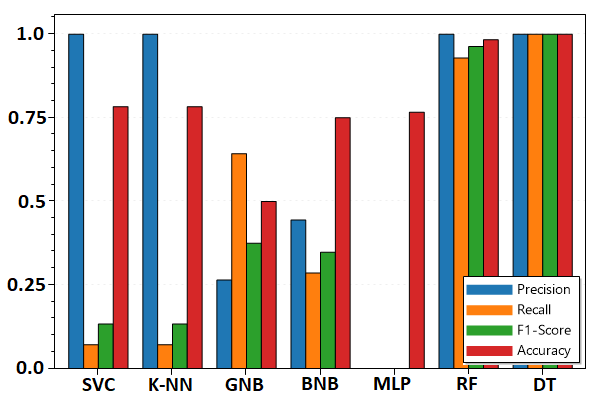}\label{disjunctive}}
  \subfigure[]{\includegraphics[height=1.6 in,width=2.3 in]{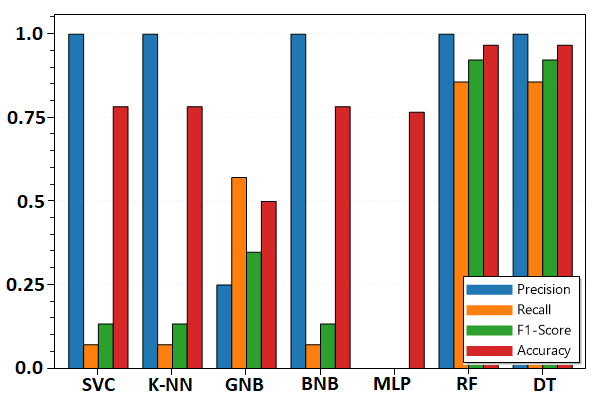}\label{conjunctive}}
  \subfigure[]{\includegraphics[height=1.6 in,width=2.3 in]{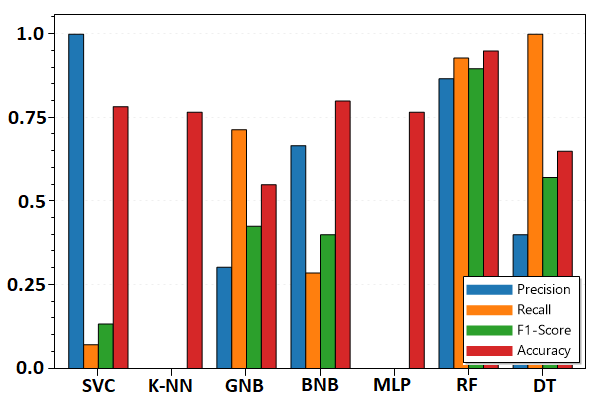}\label{cautious}}
  \caption{Precision, Recall, F1 score, Accuracy obtained based on the pignistic score from three different rules of combination (a) Disjunctive; (b) Conjunctive; (c) Cautious Conjunctive; tested with UC1, 30~sec polling interval and 10 polled outstations.}
  \vspace{-4mm}
\end{figure*}

Disjunctive rules of combination requires at least one reliable evidence. Conjunctive rules perform well if the evidences are independent and reliable. Cautious combination rules perform better when the cardinality of the frame of discernment is high. 
Hence, the comparison of conjunctive, disjunctive, and cautious conjunctive fusion technique is performed. Figs.~\ref{disjunctive},~\ref{conjunctive},~\ref{cautious} show the impact on the precision, recall, F1-score, and accuracy scores computed with the pignistic function for three different rules of combination. In the experiment, the intruder compromises the substation network, hence both the substation router and the DNP3 outstation are compromised. Among the three sensors, with the assistance of one source considered to be secure, i.e., the DNP3 master, the disjunctive fusion with DT and RF classifiers performs better in comparison to conjunctive and its cautious counterpart. 


\subsection{Comparison of Two Architectures}
\begin{figure*}[h]
  \centering
  \subfigure[]{\includegraphics[height=1.2 in,width=1.75 in]{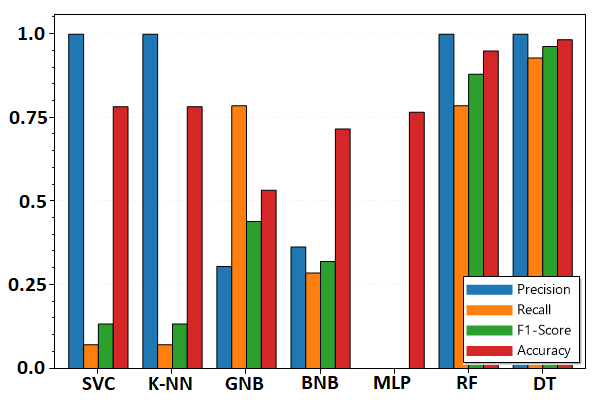}\label{uc1_id}}
  \subfigure[]{\includegraphics[height=1.2 in,width=1.75 in]{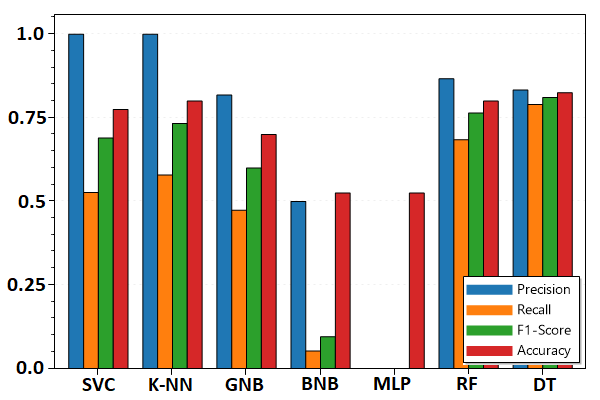}\label{uc2_id}}
  \subfigure[]{\includegraphics[height=1.2 in,width=1.75 in]{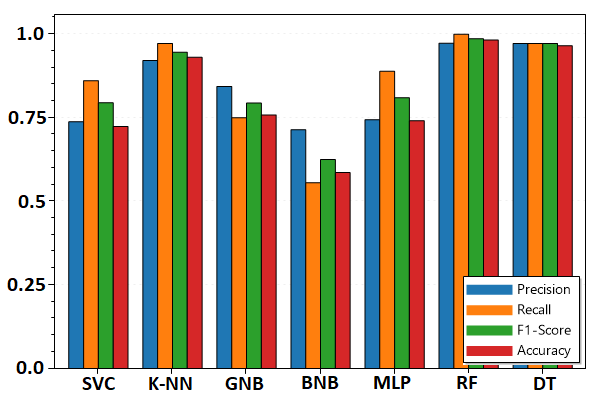}\label{uc3_id}}
  \subfigure[]{\includegraphics[height=1.2 in,width=1.75 in]{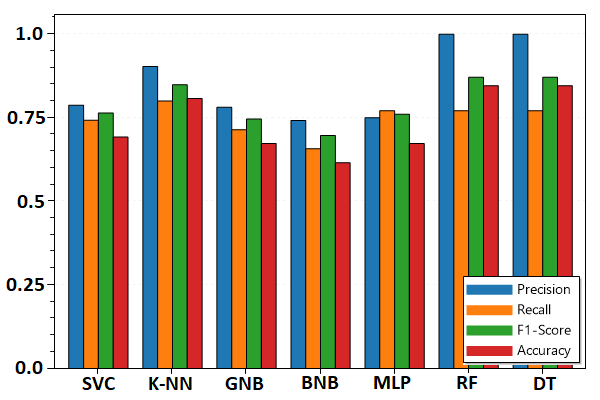}\label{uc4_id}}
  \caption{Precision, Recall, F1 score, Accuracy obtained based on the pignistic score from disjunctive based DS fusion by location and domain performed from probability scores from different classifiers with pure cyber and pure physical features combined using \textbf{FLD architecture} in Fig.~\ref{fig:two_architecture} for (a) UC1; (b) UC2; (c) UC3; (d) UC4 with 30~sec polling interval and 10 polled outstations.}
  \vspace{-4mm}
\end{figure*}

Feature-based fusion is performed prior to fusion by location (FL). If the features from diverse domains are unable to be fused due to lack of performance or due to lack of evidence from any one domain, one needs to adopt the FLD architecture, where fusion-by-location on the raw domain-specific features are performed, prior to fusion-by-domain.
 Figs. \ref{uc1}, \ref{uc2},\ref{uc3}, \ref{uc4} show the results for the FL architecture, while Figs.\ref{uc1_id}, \ref{uc2_id}, \ref{uc3_id}, \ref{uc4_id} show results from the FLD architecture. 
FLD-based fusion outperforms FL-based fusion in many scenarios, but in some cases there was not much influence. 
Hence, both may be adopted depending on the scenario.

\subsection{Impact of Time Resolution while Merging by Location}
\begin{figure*}[h]
  \centering
  \subfigure[]{\includegraphics[height=1.2 in,width=1.75 in]{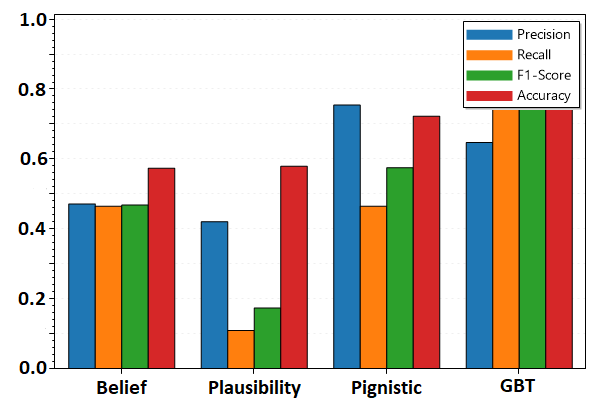}\label{dt_5s}}
  \subfigure[]{\includegraphics[height=1.2 in,width=1.75 in]{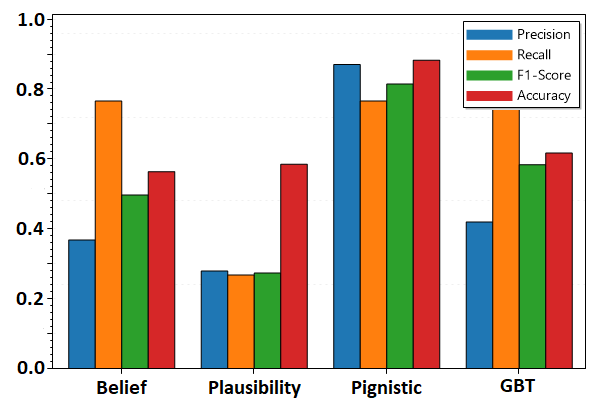}\label{dt_10s}}
  \subfigure[]{\includegraphics[height=1.2 in,width=1.75 in]{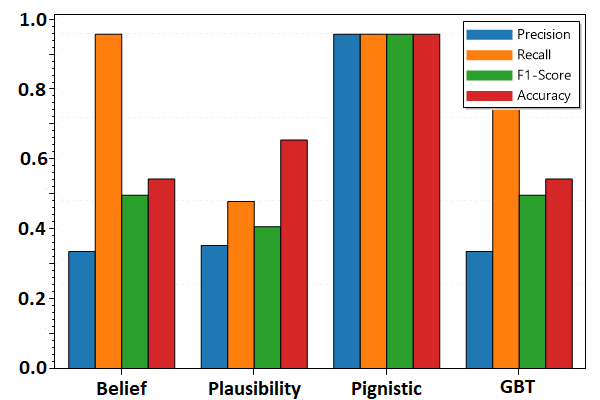}\label{dt_15s}}
  \subfigure[]{\includegraphics[height=1.2 in,width=1.75 in]{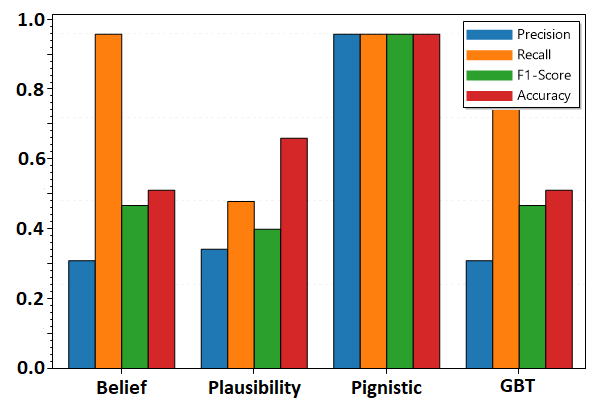}\label{dt_20s}}
  \caption{Precision, Recall, F1 score, Accuracy obtained based on the different decision criteria from disjunctive based DS fusion by location from probability scores of Decision Tree classifiers with combined cyber-physical features and different time resolutions (a) $res$ =5s; (b) $res$ =10s; (c) $res$ =15s; (d) $res$ =20s}
\end{figure*}

Since low time resolution results in noise in intrusion detection, it is advisable to consider smoothening techniques.
The physical sensor and cyber sensor time intervals between samples may vary; hence, it is essential for fusion to bring the samples to the same time frame.
This comparison evaluates detection performance based on varying time resolution, considered during time synchronization prior to fusion by location. Figs.~\ref{dt_5s},~\ref{dt_10s},~\ref{dt_15s},~\ref{dt_20s}, show the effect of different resolutions $res$ in the \textit{Mean Value based Time Synchronization} block, implemented for the probability scores obtained from the DT classifier for $UC1\;10\;OS\;30$. 
Results show that increasing the sample time lead to better decision scores, except the GBT score.  The GBT and the Belief scores are same for all the time resolution except for the $res=5s$. 

\subsection{Comparison with NSGA-2 based Feature Selection}
\begin{figure}[h]
\centering
  \includegraphics[width=0.9\linewidth]{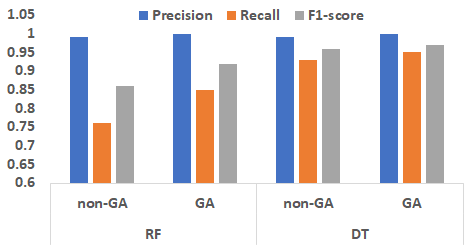}
  \caption{Comparison of the RF and DT based classifier with and without the use of GA-based feature selection.}
  \label{fig:perf_ga}
  \vspace{-4mm}
\end{figure}

Detection performance considering feature selection using the NSGA-2 algorithm varies based on the selection of type of classifier. For all the classifiers, the results obtained with the GA algorithm improve the detection performance. 
The comparison of the results for RT and DF classifiers with and without GA-based feature selection is shown in Fig.~\ref{fig:perf_ga}.

\subsection{Computation Analysis with Varying Number of Sensors}
The DS rules of combination are dependent upon 
the size of the frame of discernment as well as the number of mass functions being combined~\cite{time_complexity_combine}. 
The time complexity involved in fusion algorithms varies based on the number of evidences to fuse. It has been proven that as the set size of the frame of discernment increases, the cautious combination rule is found to be more effective~\cite{ds_algo}.




%% file: Contents/8_application.tex
\section{DSTE Evaluation Framework}
\begin{figure*}[h]
  \centering
  \includegraphics[width=1.0\linewidth]{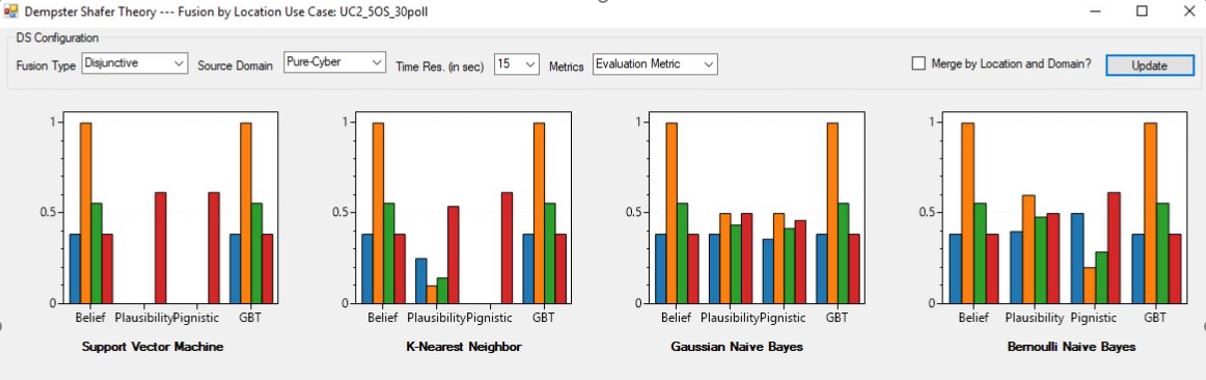}
  \caption{IDEA-I application for evaluating DSTE components for different scenarios, classifiers and fusion architectures.}
  \label{fig:ds_app}
\end{figure*}

A desktop application for IDEA-I is developed for the evaluation of DSTE rules of combination for different use-cases and parameters. Fig.~\ref{fig:ds_app} shows the application, visualizing the decision metric for $UC2\_5OS\_30poll$ for disjunctive fusion, with only cyber features, and using a mean value based time synchronization $res$ of 15~sec. The check-box labeled \textit{``Merge by Location and Domain?"} is used to select either the $FL$- or $FLD$-based architecture. The code for the DSTE evaluation application is available in Github~\cite{sahu_dsfusion}. The datasets for the evaluation of the IDEA-I are also publicly available at IEEE Dataport~\cite{fusion_dataset}.

%% file: Contents/9_conclusion.tex

\section{Conclusion}\label{sec:conclusion}
An evidence theoretic based data fusion framework for detecting cyber intrusion in power systems is presented. The framework is evaluated by studying the performance of different classifiers using DS rules of combination. Results show the evidence from the Decision Tree and Random Forest classifiers to be the best among other techniques. Results also show that higher time resolution in mean-value based time synchronization improves the decision metrics. The \textit{Pignistic Function} decision criteria is observed to be the best among all the others for all the use-cases. The FLD (autonomous architecture) outperforms the FL (centralized architecture) based fusion in many scenarios, but in some there is not much influence, so both techniques may be considered depending on the scenarios. Among the different rules of combination, the \textit{disjunctive rules} performed the best when considered with \textit{Decision Tree} and \textit{Random Forest} probability scores. Finally, an application has been developed and presented that performs these analyses and facilitates the DS theoretic framework for the fusion of cyber and physical sensors in power systems. The application is created, used, and made available to evaluate the impact of fusion type, use-case, time-resolution and architecture.